\documentclass[10pt,twocolumn,letterpaper]{article}

\usepackage[pagenumbers]{cvpr} %

\definecolor{cvprblue}{rgb}{0.21,0.49,0.74}
\newcommand{\up}[1]{\textbf{#1}$\uparrow$}
\newcommand{\down}[1]{\textbf{#1}$\downarrow$}
\newcommand{\best}[1]{\cellcolor{violet!25}\textbf{#1}}
\usepackage[pagebackref,breaklinks,colorlinks,allcolors=cvprblue]{hyperref}

\usepackage{comment}
\usepackage{multirow}
\usepackage{adjustbox}
\usepackage{colortbl}

\def\paperID{34} %
\def\confName{CVPR}
\def\confYear{2025}

\DeclareMathOperator{\diag}{diag}

\title{On the Skinning of Gaussian Avatars}

\author{Nikolaos Zioulis, Nikolaos Kotarelas, Georgios Albanis, Spyridon Thermos, Anargyros Chatzitofis\\
\href{www.moverse.ai}{Moverse}\\
{\tt\small nick@moverse.ai}
}

\begin{document}

\twocolumn[{%
\renewcommand\twocolumn[1][]{#1}%
\maketitle
\begin{center}
    \centering
    \includegraphics[width=\textwidth]{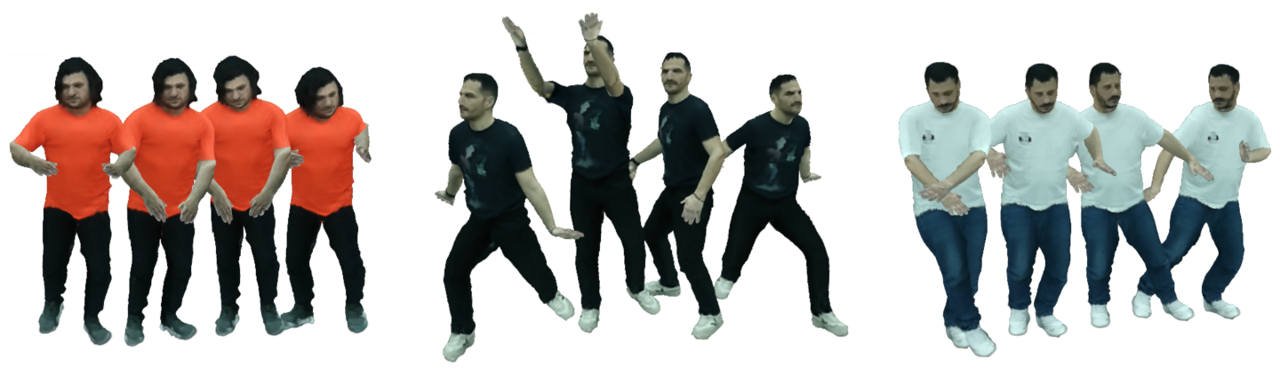}
    \captionof{figure}{
        In-the-wild skinned Gaussian avatars. Each user was captured performing a 360 turn in an A-pose, and animated using random animations from AIST++\cite{52306}. The avatars are Gaussian only, using no neural networks, and were rendered in Unity3D at over $600$ FPS.
    }
   \label{fig:teaser}
\end{center}
}]

\begin{abstract}
Radiance field-based methods have recently been used to reconstruct human avatars, showing that we can significantly downscale the systems needed for creating animated human avatars.
Although this progress has been initiated by neural radiance fields, their slow rendering and backward mapping from the observation space to the canonical space have been the main challenges.
With Gaussian splatting overcoming both challenges, a new family of approaches has emerged that are faster to train and render, while also straightforward to implement using forward skinning from the canonical to the observation space.
However, the linear blend skinning required for the deformation of the Gaussians does not provide valid results for their non-linear rotation properties. To address such artifacts, recent works use mesh properties to rotate the non-linear Gaussian properties or train models to predict corrective offsets.
Instead, we propose a weighted rotation blending approach that leverages quaternion averaging.
This leads to simpler vertex-based Gaussians that can be efficiently animated and integrated in any engine by only modifying the linear blend skinning technique, and using any Gaussian rasterizer.
\end{abstract}
    
\section{Introduction}
\label{sec:intro}

The digitization of humans as avatars has received significant attention the last years as the consolidated advances in several tasks have started to produce higher quality results. 
It lies at the intersection of computer vision and computer graphics as apart from reconstructing the appearance and geometry of the captured subjects, it also needs to simulate them via animation, deformation, and rendering.
The emergence of modern view synthesis technologies aligned with a mature state-of-the-art in parametric body modeling \cite{loper2015smpl,osman2020star,Pavlakos_2019_CVPR}, color-based motion capture \cite{choutas2020monocular,goel2023humans,zhang2023pymaf}, and human segmentation \cite{ke2022modnet} that provided consistent and multimodal priors and constraints.
What was previously attainable with systems that needed more than 100 cameras \cite{collet2015high} has been downscaled to much sparser setups \cite{zhao2022humannerf,dong2024gaussian,niu2024bundle,luo2025deblur} and even a single mobile phone \cite{weng2022humannerf,hu2024gaussianavatar}.

Neural radiance fields (NeRFs) \cite{mildenhall2021nerf} were first applied to avatar reconstruction using volume rendering.
Early approaches deformed camera ray samples using the closest bone transforms \cite{su2021nerf} or via inverse skinning of the closest surface point \cite{wang2022arah} when using parametric body models.
Using such a canonical space is a requirement for volume rendering and ray integration, but proved challenging, as the backward mapping from the observation space to the canonical one is ill-posed and produces multiple solutions.
Although improved \cite{chen2021snarf} and faster \cite{chen2023fast} methods surfaced, the volume rendering itself is highly inefficient. 
To improve efficiency a parallel line of work relied on implicit neural texture fields, defined on the unwrapped texture space of a mesh representation.

The introduction of Gaussian Splatting (GS) \cite{kerbl20233d} changed the field, as its differentiable rasterization solved both aforementioned problems simultaneously.
Not only was rasterization much more efficient than ray-based integration, but it also offered an explicit point--based representation that could be forward mapped to the observation space effectively, removing the challenges associated with the backward mapping between the observation and canonical space.
Indeed, early Gaussian avatar works presented fast training times and faster rendering rates \cite{li2023human101,qian20243dgs}. 

The common denominator of all recent Gaussian avatar works is the linear blend skinning (LBS) of the Gaussians which allows them to deform from a canonical representation to the observation space and thus be optimized.
However, a Gaussian point representation is a complex representation of a rotated and scaled ellipsoid, accompanied by spherical harmonics coefficients (SH) and an opacity value, used for alpha blending.
Ignoring the rotation invariant properties, namely the scale and opacity, LBS can deform the positions of the scaled Gaussians, but cannot properly rotate the SH coefficients.
Further, linearly blended transforms do not produce valid rotations, and although highly used in prior work, they do not consistently rotate the Gaussians, even when converted to other rotation representations, or ortho-normalized to a valid rotation matrix.
To overcome this, earlier works heavily relied on neural networks to predict  the Gaussian rotation \cite{hu2024gaussianavatar,pang2024ash,kocabas2024hugs,moreau2024human,li2023animatable}, or predict rotation offsets to correct the LBS rotations \cite{zielonka2023drivable,li2023human101,qian20243dgs}.
More recent works employ the rotation information \cite{shao2024splattingavatar,paudel2024ihuman} (\textit{e.g.}~face normal direction) embedded in the mesh structure and rotate the Gaussian's accordingly.
But these result in either the loss of rotation degrees of freedom, suffer from interaction with the position updates or introduce complexity when animating and rendering.

Regarding view dependent effects, as enabled by the use of spherical harmonics, most works on LBS-based Gaussian avatar works only predict the constant zero-order coefficients or color directly \cite{moreau2024human,shao2024splattingavatar,pang2024ash,zielonka2023drivable,hu2024gaussianavatar}, or use neural networks to predict higher order coefficients \cite{kocabas2024hugs}.
Still, the models would need to be conditioned on pose to be able to learn the necessary animation invariance required to properly model the SH consistently across poses.
To address this, a group of prior works have been transforming the view direction for each Gaussian using the rotation of the LBS transform \cite{jung2023deformable,qian20243dgs,liu2024animatable}.
Still, as aforementioned, the LBS rotation is not proper, leading to inconsistently transformed directions.

\begin{figure}[t]
    \centering
    \includegraphics[width=\columnwidth]{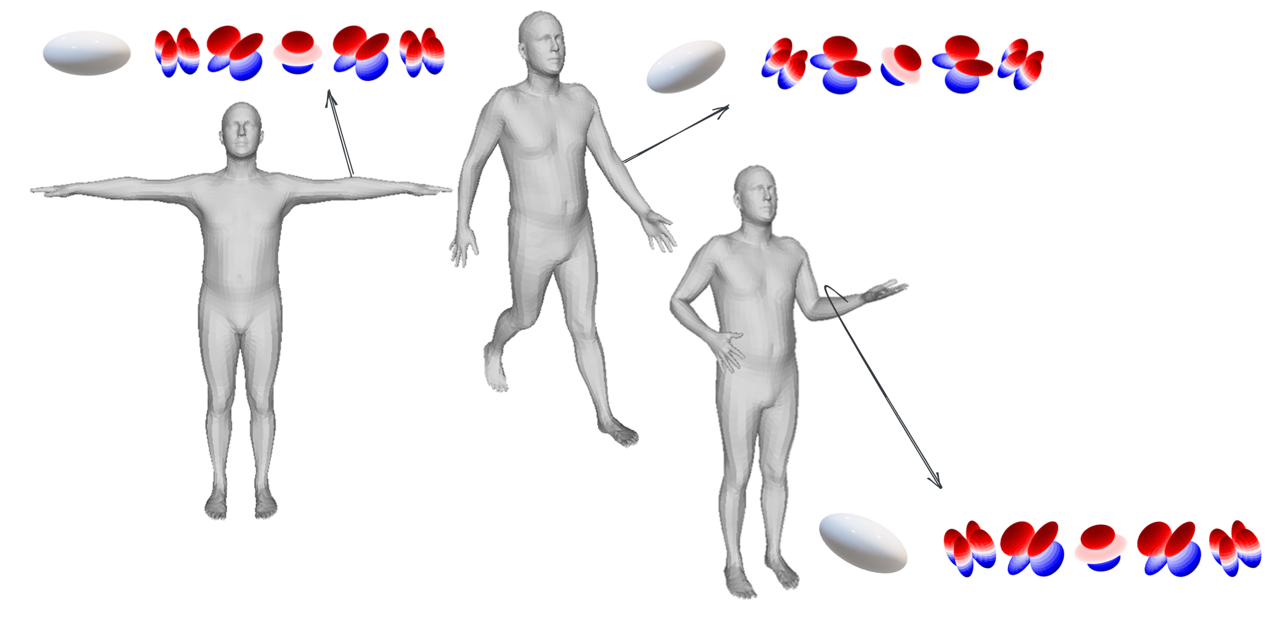}
    \caption{Skinning Gaussians relies on linear blending to transform the positions. However, the ellipsoid's rotation and spherical harmonics coefficients need to be properly rotated as well.}
    \label{fig:sh_rotation}
\end{figure}

In this work we present complete skinning for Gaussians avatars, integrating a number of techniques to animate all Gaussian properties according to the avatar's skinning, while preserving rotational consistency when deforming the avatar's Gaussians, as seen in Figure~\ref{fig:sh_rotation}. 
To achieve this, we employ quaternion averaging to perform weighted rotation blending using the Gaussians' skinning weights, and then use that blended rotation to rotate the SH coefficients.
The resulting method essentially employs a vertex-based Gaussian animation technique, one that is highly parallel, agnostic to the Gaussian rasterizer used, and does not require modifications to the rasterizer itself.

\section{Related Work}
\label{sec:related}

Human avatar reconstruction aims to digitize the captured subjects to a simulation-ready digital appearance representation.
As simulation covers a wide-spectrum of capabilities, in this work we follow recent literature and focus on animation as the core simulation property, enabling the avatar to deform with human joint motion.
Similarly, we consider freely moving captured subjects, relaxing the capturing conditions away from fixed pose -- \textit{i.e.}~stationary -- captures.
Therefore, we review related works in this specific scope, and will not be discussing non-animation-ready literature \cite{newcombe2015dynamicfusion,collet2015high,saito2020pifuhd}, or those not capturing appearance \cite{yu2018doublefusion}.

\subsection{Earlier Works}
While prior similar works for head avatars existed beforehand \cite{ichim2015dynamic}, VideoAvatars \cite{alldieck2018video} was one of the first works to use a parametric (i.e. simulation-ready) human body model to constrain the reconstruction process using silhouettes and deliver avatars from a single video.
Follow-up works either relied on RGB-D inputs \cite{zuo2020sparsefusion,zhi2020texmesh} to improve geometric fidelity, or expanded the constraints using data-driven outputs like keypoints \cite{zuo2020sparsefusion,wenninger2020realistic}, normal maps\cite{zhi2020texmesh}, segmentation \cite{alldieck2018detailed}, and/or improved silhouettes \cite{natsume2019siclope}.
To improve appearance reconstruction without compromising geometric fidelity advanced differentiable rendering techniques started being integrated into these methods \cite{zhi2020texmesh,habermann2021real}.

\subsection{Radiance Fields}
Following the advent of neural radiance fields \cite{mildenhall2021nerf}, new techniques emerged that employed volume rendering and implicit neural networks. 
Although some used a simple skeleton \cite{su2021nerf} or point cloud \cite{peng2021neural} representation, most used LBS to deform a canonical template mesh and implicit fields defined either in texture space \cite{liu2021neural}, via canonical space coordinates \cite{xu2021h,chen2021animatable}, blendshape weights \cite{peng2021animatable}, or via hybrid schemes involving the color images as well \cite{wang2022nerfcap}.
Crucially, rays were stratified sampled and deformed to the canonical space to allow for the optimization of the NeRF avatars.
This backwards deformation is challenging, produces non-unique solutions, is biased towards the poses available when fitting the neural field, and is also slow when considering that it was performed for all ray samples.
As a result improved techniques for differential forward skinning surfaced, improving both quality \cite{chen2021snarf} and speed \cite{chen2023fast}.

\subsection{Gaussian Splatting}
The GS technique was quickly adopted to overcome the performance issues of volume rendering, and additionally offered a side-advantage; support for forward mapping via LBS skinning instead of inverse backward mapping.
Many variants were proposed, the majority exploiting parametric body mesh representations and LBS to warp the Gaussians to the observation space.
Human101 \cite{li2023human101}, HUGS \cite{kocabas2024hugs} HuGS \cite{moreau2024human}, D3GA \cite{zielonka2023drivable} and SplatArmor \cite{jena2023splatarmor} learned Gaussian splat offers via neural networks.
ASH \cite{pang2024ash}, AnimateableGaussians \cite{li2023animatable} and GaussianAvatar \cite{hu2024gaussianavatar} used texture-based autoencoders to learn the Gaussian splat properties.
The vast majority of avatar reconstruction using Gaussian splats \cite{fei20243d,wang2024survey} employ  neural networks to either predict the splats' properties directly or offsets to those deformed by LBS.
This helps offset the weakness of LBS to properly handle the splat properties that rotate, the ellipsoid's rotation and the view dependent spherical harmonics coefficients.
The reason is that linearly blended rotations are not valid rotations for a good portion of typical vertex skinning weights.

For the splat rotations a recent direction has been to use the parametric body mesh's faces.
iHuman \cite{paudel2024ihuman} and SAGA \cite{chen2024saga} attach the splats' rotation on the face, while SplattingAvatar \cite{shao2024splattingavatar} uses the tangent--bitangent--normal rotation composition to rotate the splats' rotation component.
While the former loses rotational degrees--of--freedom, the latter introduces complexity and also couples the splat position and rotation properties.
In contrast, our skinned gaussian avatars allow for the proper -- and full -- rotation of the gaussian splats using the mesh's skinning weights.

Another property that suffers due to LBS is the view dependency, as provided by the spherical harmonics coefficients. 
While some works ignore orders above zero, those that do not need to account for the avatar's deformation, and thus, rotation of the spherical harmonic coefficients.
One solution employed by Human101 \cite{li2023human101} is to rotate the first order coefficients using the LBS rotation.
For the other coefficients, the Wigner D-matrix can be used, but prior work that used it found no benefits \cite{hu2024gauhuman}.
Another approach is to deform the view direction using each splat's LBS transform as in \cite{jung2023deformable} and 3D-GS Avatar \cite{qian20243dgs}, essentially canonicalizing it. 
This rotation of the viewing direction was also implemented directly in the rasterizer in Animatable 3D Gaussians \cite{liu2024animatable} along with LBS itself for the positions and rotations.
However, both the Wigner D-matrix and view direction canonicalization approaches suffer from the improper LBS rotation matrices, which manifest as inconsistent rotations.
We propose using a properly skinned rotation, so that all the SH coefficients can also be rotated accordingly.

\newcommand{\V}{\mathbf{V}}
\newcommand{\F}{\mathbf{F}}
\newcommand{\pose}{\boldsymbol{\theta}}
\newcommand{\sh}{\boldsymbol{s}}
\newcommand{\opacity}{\alpha}
\newcommand{\scaling}{\boldsymbol{\sigma}}
\newcommand{\rotationq}{\boldsymbol{q}}
\newcommand{\position}{\boldsymbol{\mu}}
\newcommand{\gaussian}{\{\position, \rotationq, \scaling, \opacity, \sh \}}
\newcommand{\gaussiani}{\{\position_i, \rotationq_i, \scaling_i, \opacity_i, \sh_i \}}
\newcommand{\covariance}{\boldsymbol{\Sigma}}
\newcommand{\rotation}{\boldsymbol{R}}
\newcommand{\scale}{\boldsymbol{S}}
\newcommand{\jacobian}{\boldsymbol{J}}
\newcommand{\view}{\boldsymbol{V}}
\newcommand{\pixel}{\boldsymbol{c}}
\newcommand{\skinning}{\boldsymbol{w}}
\newcommand{\transform}{\boldsymbol{T}}
\newcommand{\translation}{\boldsymbol{t}}

\section{Method}
\begin{figure*}
    \centering
    \includegraphics[width=\textwidth]{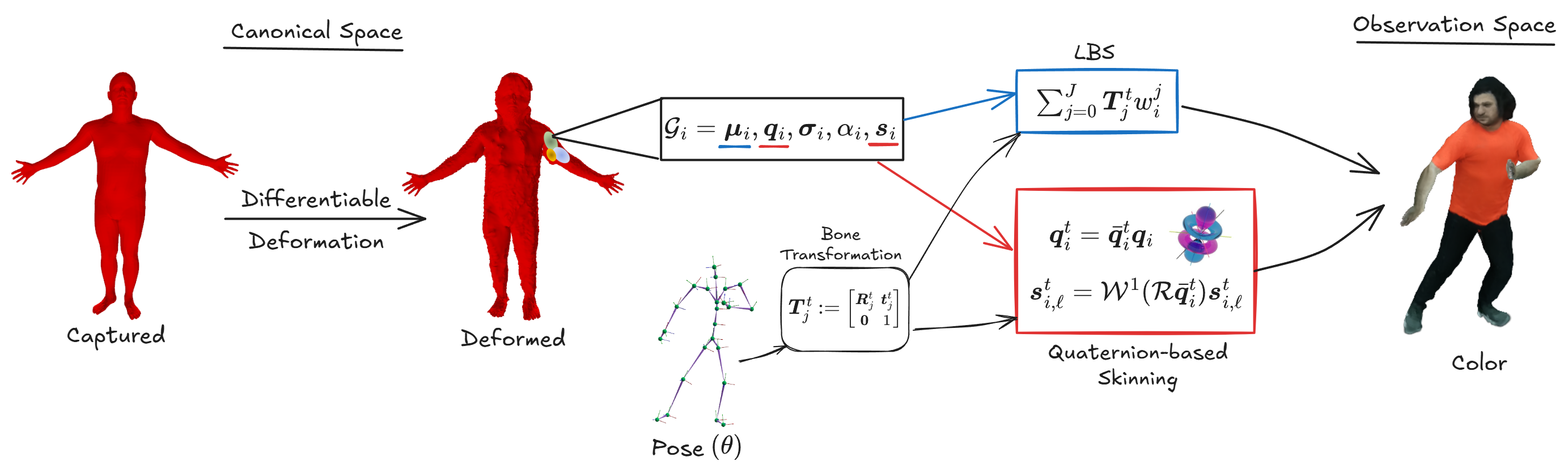}
    \caption{Method overview: A preprocessing step deforms of the captured shape using silhouette information (left); the rotation-related parameters of the Gaussians are skinned in the quaternion space (middle); the human body pose transformation is used to weight the skinning process (middle); the skinning result along with the scale and opacity parameters lead to the final animated 3D avatar (right). }
    \label{fig:method}
\end{figure*}

We first (\S\ref{method:preliminiaries}) give an overview of LBS-based Gaussian Avatars by outlining the GS technique and then how it is adapted to capture animated human avatars.
This sets the stage for describing (\S\ref{method:skinned}) how we properly skin all Gaussian components when animating the avatar.

\subsection{Preliminaries}
\label{method:preliminiaries}
GS \cite{kerbl20233d} represents scenes with a set of $N$ unorganized 3D Gaussians $\{\mathcal{G}_{i} \mid i \in [\![1,N]\!] \}$, where each one is defined as:

\begin{equation}
    \mathcal{G}_{i} = \gaussiani,
\end{equation}

where $\position \in \mathbb{R}^3$ is the position, $\rotationq \in \mathbf{H}$ is its rotation represented as a quaternion, $\scaling \in \mathbb{R}^3$ is its scale, $\opacity \in (0, 1)$ is its opacity, and $\sh \in \mathbb{R}^{16 \times 3}$ are third order spherical harmonics coefficients for each color channel.
In practice, some of these are transformed versions of the actual parameters of the Gaussian, specifically:
  
\begin{itemize}[leftmargin=1.25in]
    \item $\scaling = e^{\scaling^g}$,
    \item $\rotationq = \rotationq^g / || \rotationq^g ||_2, \rotationq^g \in \mathbb{R}^4$, 
    \item $\opacity = \varsigma(\opacity^g), \opacity^g \in \mathbb{R}$,
\end{itemize}

\noindent where the superscript $g$ denotes the actual parameters, and $\varsigma$ is the sigmoid function. The rotation and scaling properties define the Gaussian's covariance:

\begin{equation}
    \covariance = \rotation\scale\scale^{T}\rotation^{T},
\end{equation}

\noindent when given in their matrix forms $\scale := \diag(\scaling)$ and $\rotation := \mathcal{R}(\rotationq) \in \mathbb{SO}^3$, with $\mathcal{R}$ representing the quaternion to rotation matrix conversion function.

Using the covariance, its Gaussian's spatial extend is defined using its spatially varying opacity:

\begin{equation}
    \varrho(\mathbf{x}) = \opacity e^{\frac{1}{2}(\mathbf{x} - \position)^{T}\covariance^{-1}(\mathbf{x} - \position)}.
\end{equation}

\noindent The Gaussians are splatted using the Elliptical Weighted Average (EWA) technique \cite{zwicker2002ewa} after projecting them to an image using a view transform matrix $\view$.
Their associated image space covariance is derived using the Jacobian of the project matrix $\jacobian$:

\begin{equation}
    \covariance^{2D} = \jacobian\view\covariance\view^{T}\jacobian^{T}
\end{equation}

The resulting color is accumulated after depth sorting the splatted Gaussians for each pixel, keeping M splat contributions per pixel, and accumulating them using alpha blending in a volume-like integration manner:

\begin{equation}
  \pixel = \sum_{p=0}^{M} \pixel_p\opacity_p\prod_{q=1}^{p-1}(1-\opacity_q).
\end{equation}

However, this representation is for static scenes, so when considering deformed -- or, animated -- avatars, the Gaussians need to be likewise animated. 
Using a parametric body model as a canonical representation \cite{loper2015smpl,Pavlakos_2019_CVPR}, each Gaussian is initially defined in the canonical frame, and accompanied by a skinning weights vector $\skinning_i \in \mathbb{R}^J, \mbox{\textit{s.t.}} \, w_{i}^{j} \ge 0 \, \mbox{\textit{and}} \, \sum_{j=0}^{J}w_{i}^{j} = 1$, where $J$ is the number of joints describing the body's articulation.
Using a specific pose at time $t$, and via a forward kinematics, we derive the body's bones transformations set $\{\transform_j^t := \left[ \begin{smallmatrix} \rotation_j^t & \translation_j^t\\ \mathbf{0}&1 \end{smallmatrix} \right] \in \mathbb{SE}(3) , j \in [\![1..J]\!] \}$.
The weights $\skinning_{i}$ provide an association of each Gaussian with the bones, and its positions are deformed using linear transform blending:
\begin{equation}
\label{lbs_position}
    \position_{i}^{t} = \mathcal{T}_{i}^t\position_{i}, \, \mathcal{T}_{i}^{t}=\sum_{j=0}^{J}\transform_{j}^{t}w_{i}^{j},
\end{equation}
with $\mathcal{T}_{i}^{t}$ being the linearly blended transforms using the skinning weights $\skinning_{i}$.

The LBS transform $\mathcal{T}$ is a standard technique for animating the positions, although there exists prior work that explores optimizing the skinning weights \cite{liao2024vinecs} to improve realism.
It is also typically used to rotate the Gaussian's rotation component \cite{qian20243dgs,hu2024gauhuman,moreau2024human,li2023animatable,liu2024animatable} and in some cases the first order SH coefficients \cite{hu2024gauhuman}, and/or transform the viewing direction to the canonical coordinate system \cite{jung2023deformable,qian20243dgs,liu2024animatable}.
But the problem is that linearly blending rotations does not produce a valid rotation for a large portion of typical skinning weights \cite{kavan2007skinning}.
While this results into volume loss in certain configurations for the positions, the effect is more pronounced when only considering the rotation component, making it inadequate for applying it to the rotation-aware Gaussian components, $\rotationq$ and $\sh$.

\subsection{Skinned Gaussians}
\label{method:skinned}

It is clear that in order to properly skin the Gaussians, we need them to properly blend the rotation components of bone transforms.
One solution to this problem is weighted rotation averaging, and specifically quaternion averaging \cite{markley2007averaging}.
With $\mathcal{R}^{-1}$ being the rotation matrix to quaternion conversion function, we extract the rotational component of the bone transforms $\transform_{j}^{t}$ as a quaternion $\mathbf{b}_{j}^{t} = \mathcal{R}^{-1}(\rotation_{j}^{t})$.
To calculate the weighted rotation average for each Gaussian $\mathcal{G}_{i}$ we need to construct the matrix $\mathbf{A}$:
\begin{equation}
\label{average_quaternion}
    \mathbf{A}_{i}^{t} = \sum_{j=0}^{J} w_{i} ^{j} (\mathbf{b}_{j}^{t})^{T}\mathbf{b}_{j}^{t}.
\end{equation}
The weighted average quaternion $\bar{\rotationq}_{i}^{t}$ is the eigenvector corresponding to the maximum eigenvalue of $\mathbf{A}$.
Then for each pose $t$ we can properly skin and rotate each Gaussians $\mathcal{G}_i$ rotation component:
\begin{equation}
\label{blended_rotation}
    \rotationq_{i}^{t} = \bar{\rotationq}_{i}^{t} \rotationq_{i}.
\end{equation}

Having obtained a properly skinned rotation, it can also be used to rotate the SH coefficients. 
This can be achieved with the Wigner D-matrices $\mathcal{W}^{l}(\mathbf{R})$ \cite{wigner1931gruppentheorie,gilmore2008lie,aubert2013alternative} which map elements $\mathbf{R}$ of $\mathbb{SO}^3$ to $(2l + 1) \times (2l + 1)$ matrices.
These can be used to rotate each SH order's $l$ vector of coefficients:
\begin{equation}
\label{sh_rotation}
\begin{gathered}
    \sh_{i,0:1}^{t} = \sh_{i,0:1}^{t},\\
    \sh_{i,1:4}^{t} = \mathcal{W}^{1}(\bar{\rotation}_{i}^{t}) \sh_{i,1:4}^{t},\\
    \sh_{i,4:9}^{t} = \mathcal{W}^{2}(\bar{\rotation}_{i}^{t}) \sh_{i,4:9}^{t},\\
    \sh_{i,9:16}^{t} = \mathcal{W}^{3}(\bar{\rotation}_{i}^{t}) \sh_{i,9:16}^{t},
\end{gathered}
\end{equation}
where the performed rotation $\bar{\rotation} = \mathcal{R}(\bar{\rotationq})$ is the skinned one, and the SH coefficients $\sh$ are sliced to each respective order $l \in {0..3}$.
Prior works have either only used the zeroth order SH coefficients, therefore ignoring view dependent effects, or only rotated the first order coefficients or the viewing direction with the improper LBS rotation.

Taking into account the the scale $\scaling$ and opacity $\opacity$ are rotation invariant properties, the remaining Gaussian components, position $\position$, rotation $\rotationq$, and SH coefficients $\sh$, are now properly skinned, with the weights $\skinning$ for an articulated body of $J$ joints, using Eq.\eqref{lbs_position}, \eqref{blended_rotation} and \eqref{sh_rotation} respectively.

\section{Experiments}

Since skinning the Gaussians is a horizontal technique that can be applied to most prior Gaussian Avatar works, our evaluation is multi-faceted.
Our goal is to cover multiple techniques where proper skinning can replace traditional LBS (\S\ref{sec:lbs_comparison}), or replace other rotation estimation techniques (\S\ref{sec:mesh_comparison}), and also replace view direction canonicalization (\S\ref{sec:viewdir_comparison}).
Finally, we implement a pure Gaussian Avatar baseline (\S\ref{sec:sga_comparison}), one that is not using any neural network and instead only optimizes skinned Gaussians.
For this implementation, we show in-the-wild results and demonstrate real-time performance by integrating them in a game engine (\S\ref{sec:wild_results}).
Aiming to demonstrate generality, and apart from the data we recorded, our experiments are performed across 4 different datasets, each with different traits, and using three different body models.
For all integrations with other methods, we use the publicly available implementations of the authors and also use roma \cite{bregier2021deepregression} and e3nn \cite{lapchevskyi2020euclidean} for our Gaussian skinning implementation.
We mainly evaluate visual quality using PSNR, SSIM, and LPIPS \cite{zhang2018unreasonable} metrics.

\subsection{Comparison to linear blending}
\label{sec:lbs_comparison}
For our first experiment, we integrate Gaussian skinning into HUGS \cite{kocabas2024hugs}, a method that uses a triplane feature extractor and a set of MLPs to predict position offsets, the remaining Gaussian components, including third-order SH coefficients, as well as skinning weights for a canonical SMPL model.
We adapt it by converting the learned rotation -- parameterized as 6D \cite{zhou2019continuity} -- to a quaternion and then applying our rotation skinning via Eq.~\eqref{blended_rotation}.
In addition, the SH coefficients are also skinned using Eq.~\eqref{sh_rotation}.
Table~\ref{tab:hugs} presents the results of the original implementation and our modified one using our Gaussian skinning method on the NeuMan dataset \cite{jiang2022neuman}.
As we only modify the skinning component, we report the results for the human avatar only, where it is evident that properly skinning all Gaussian components offers a consistent image quality performance boost.

\begin{table}[!htbp]
\centering
\resizebox{0.85\columnwidth}{!}{%
\begin{tabular}{ccccc}
Sequence & HUGS \cite{kocabas2024hugs} & \up{PSNR} & \up{SSIM} & \down{LPIPS} \\ \hline
\multirow{2}{*}{Lab} & Original & 22.07 & 0.7952 & 0.1769 \\
 & Skinned & \best{22.38} & \best{0.7969} & \best{0.1670} \\ \hline
\multirow{2}{*}{Bike} & Original & 21.51 & 0.7696 & 0.1841 \\
 & Skinned & \best{21.79} & \best{0.7701} & \best{0.1773} \\ \hline
\multirow{2}{*}{Seattle} & Original & 19.70 & 0.7052 & 0.1648 \\
 & Skinned & \best{19.81} & \best{0.7088} & \best{0.1503} \\ \hline
\multirow{2}{*}{Citron} & Original & 21.66 & 0.7706 & 0.1249 \\
 & Skinned & \best{22.12} & \best{0.7778} & \best{0.1199} \\ \hline
\multirow{2}{*}{Jogging} & Original & 18.48 & 0.6800 & 0.2218 \\
 & Skinned & \best{18.63} & \best{0.6816} & \best{0.2099}
\end{tabular}%
}
\caption{Results on NeuMan \cite{jiang2022neuman} using HUGS \cite{kocabas2024hugs}.
}
\label{tab:hugs}
\end{table}

\subsection{Comparison to mesh-based rotation}
\label{sec:mesh_comparison}
Next, we compare skinned Gaussians with those embedded in meshes, where the rotation is calculated by the mesh faces.
The first mesh-based method for which we integrate our Gaussian skinning is iHuman \cite{paudel2024ihuman} that uses a neural network to predict second-order SH coefficients, and only optimizes a scalar rotation angle parameter, combining it with each face's normal direction to derive the Gaussians' animation rotation.
We modify iHuman to optimize a quaternion rotation, and employ our skinning technique to calculate the animated rotation for each vertex, and then perform a quaternion average for each face, as the Gaussians lie on the face centroids.
We use this rotation to rotate the neural network predicted SH coefficients.

The second mesh-based method for which we integrate our Gaussian skinning is SplattingAvatar \cite{shao2024splattingavatar}.
Similarly to iHuman, it derives the Gaussians' animated rotation from the mesh faces using the normal-tangent-bitangent composition.
Contrary to iHuman, the Gaussians are freely positioned on the faces using barycentric coordinates, so after the initial vertex-based rotation calculation, the face quaternion averaging uses the barycentric weights.
As the original method only uses the zeroth-order SH coefficients, we increase it to the second order and rotate it using Eq.~\eqref{sh_rotation}.

Tables~\ref{tab:ihuman} and \ref{tab:skinningavatar} present the results of the original implementations and our adaptations with our proposed skinning using the PeopleSnapshot dataset \cite{alldieck2018detailed}.
Both methods benefit from properly skinning the rotations as results are improved on average across all sequences.
More importantly, as seen in Table~\ref{tab:skinningavatar}, SplattingAvatar optimizes the number of Gaussians using a walk-on-triangle scheme, and our modified variant always produces fewer splats.
This is an indicator that the optimization process itself is improved as the Gaussians are consistently rotated, and the rotation updates are not influenced directly by the position ones.

\begin{table}[!htbp]
\centering
\begin{tabular}{@{}lcccc@{}}
\toprule
Casual                         & \multicolumn{1}{c}{\cite{paudel2024ihuman}} & \up{PSNR} & \up{SSIM} & \down{LPIPS} \\ \midrule
\multirow{2}{*}{male-3}   & Original                    & 27.82     & 0.9658    & \best{0.0198}       \\
                                 & Skinned                     & \best{27.89}     & \best{0.9660}     & 0.0206       \\ \hline
\multirow{2}{*}{female-3} & Original                    & 21.99     & 0.9392    & 0.0505       \\
                                 & Skinned                     & \best{22.26}     & \best{0.9405}    & \best{0.0466}       \\ \hline
\multirow{2}{*}{male-4}   & Original                    & 25.30      & 0.9511    & 0.0369       \\
                                 & Skinned                     & \best{25.46}     & \best{0.9529}    & \best{0.0361}       \\ \hline
\multirow{2}{*}{female-4} & Original                    & 26.89     & 0.9622    & 0.0244       \\
                                 & Skinned                     & \best{26.92}     & \best{0.9624}    & \best{0.0230}        \\ %
\end{tabular}
\caption{Results on PeopleSnapshot \cite{alldieck2018detailed} (casual subset) using iHuman \cite{paudel2024ihuman}, and our adapted skinned GS version.}
\label{tab:ihuman}
\end{table}

\begin{table}[!htbp]
\centering
\resizebox{\columnwidth}{!}{%
\begin{tabular}{@{}llcccc@{}}
\toprule
Casual                    & \multicolumn{1}{c}{\cite{shao2024splattingavatar}} & \up{PSNR}    & \up{SSIM}     & \down{LPIPS}  & \multicolumn{1}{l}{\down{\# splats}} \\ \midrule
\multirow{2}{*}{male-3}   & Original                            & 29.83        & 0.9729        & 0.0710        & 67389                                \\
                          & Skinned                             & \best{29.97} & \best{0.9746} & \best{0.0704} & \best{52026}                         \\ \hline
\multirow{2}{*}{female-3} & Original                            & 28.28        & 0.9714        & \best{0.0819} & 58733                                \\
                          & Skinned                             & \best{28.44} & \best{0.9721} & 0.0825        & \best{53201}                         \\ \hline
\multirow{2}{*}{male-4}   & Original                            & 28.26        & 0.8712        & 0.0900        & 77504                                \\
                          & Skinned                             & \best{28.36} & \best{0.9724} & \best{0.0896} & \best{63057}                         \\ \hline
\multirow{2}{*}{female-4} & Original                            & 29.26        & 0.9705        & \best{0.0711} & 57092                                \\
                          & Skinned                             & \best{29.46} & \best{0.9719} & \best{0.0711} & \best{45704}                         \\ %
\end{tabular}%
}
\caption{Results on PeopleSnapshot \cite{alldieck2018detailed} (casual subset) using SplattingAvatar \cite{shao2024splattingavatar}, and our adapted skinned GS version.}
\label{tab:skinningavatar}
\end{table}

\subsection{Comparison to view direction canonicalization}
\label{sec:viewdir_comparison}
Animatable 3D Gaussians \cite{liu2024animatable} also use LBS-based skinned Gaussians, but differentiate their approach by transforming the view direction into the canonical space to evaluate the SH coefficients, a technique also employed by 3D-GS Avatar \cite{qian20243dgs}.
This is implemented by a custom Gaussian rasterizer that receives the bone transforms as input, while also using hash encoding and MLPs to predict some of the Gaussian properties.
For our comparison, we used the traditional Gaussian splat rasterizer \cite{kerbl20233d}, using our skinned Gaussians that also rotate the SH coefficients.
We present results for both the PeopleSnapshot \cite{alldieck2018detailed} dataset using the SMPL body \cite{loper2015smpl}, but also for the GalaBasketball \cite{liu2024animatable} dataset, which is a synthetic dataset using a custom rigged and skinned 3D model.
It should be noted that for a fair comparison we disable the time-dependent ambient occlusion as the traditional rasterizer does not support it.
Tables~\ref{tab:viewdir-gala} and \ref{tab:viewdir-ps} present the results for $4$ single actor scenes from the GalaBasketball dataset, and the $4$ PeopleSnapshot sequences, respectively.
Even though the approaches are conceptually similar, we observe a consistent performance boost when using the properly blended rotations.
However, our approach has another advantage, that it does not require adaptation to the rasterizer, ensuring straightforward interoperability with any GS rasterization improvements \cite{liang2024gs,liang2024analytic,yan2024multi,talegaonkar2024volumetrically}.

\begin{table}[!htbp]
\centering
\begin{tabular}{ccccc}
\toprule
Single Actor & \cite{liu2024animatable} & \textbf{\up{PSNR}} & \textbf{\up{SSIM}} & \textbf{\down{LPIPS}} \\ \hline
\multirow{2}{*}{Dribble} & Original & 37.56 & 0.9926 & 0.0046 \\
 & Skinned & \best{37.84} & \best{0.9929} & \best{0.0045} \\ \hline
\multirow{2}{*}{Idle} & Original & \best{38.72} & \best{0.9958} & 0.0034 \\
 & Skinned & 38.60 & \best{0.9958} & \best{0.0032} \\ \hline
\multirow{2}{*}{Shot} & Original & 38.95 & 0.9945 & \best{0.0043} \\
 & Skinned & \best{39.08} & \best{0.9947} & 0.0045 \\ \hline
\multirow{2}{*}{Turn} & Original & \best{37.57} & \best{0.9952} & 0.0037 \\
 & Skinned & 37.16 & \best{0.9952} & \best{0.0036}
\end{tabular}%
\caption{Results on GalaBasketball \cite{liu2024animatable} using Animatable 3D Gaussians \cite{liu2024animatable}, and our skinned GS adaptation.}
\label{tab:viewdir-gala}
\end{table}

\begin{table}[!htbp]
\centering
\begin{tabular}{lcccc}
\toprule
Casual & \cite{liu2024animatable} & \textbf{\up{PSNR}} & \textbf{\up{SSIM}} & \textbf{\down{LPIPS}} \\ \hline
\multirow{2}{*}{male-3} & Original & 29.04 &	0.9704	& 0.0264 \\
 & Skinned & \best{29.12} & \best{0.9706} & \best{0.0263} \\ \hline
\multirow{2}{*}{female-3} & Original & 24.59 & \best{0.9535} & \best{0.0398} \\
 & Skinned & \best{24.59} & \best{0.9535} & 0.0399 \\ \hline
\multirow{2}{*}{male-4} & Original & 26.14	& \best{0.9554} &	0.0490 \\
 & Skinned & \best{26.18}	& \best{0.9554} &	\best{0.0489} \\ \hline
\multirow{2}{*}{female-4} & Original & 27.28	& 0.9634	& 0.0281 \\
 & Skinned & \best{27.33} & \best{0.9637} & \best{0.0280}
\end{tabular}%
\caption{Results on PeopleSnapshot \cite{alldieck2018detailed} (casual subset) using Animatable 3D Gaussians \cite{liu2024animatable}, and our skinned GS adaptation.}
\label{tab:viewdir-ps}
\end{table}

\begin{table*}[!htbp]
\centering
\resizebox{\textwidth}{!}{%
\begin{tabular}{@{}lcccccccccccccccl@{}}
\toprule
Subject                    & \multicolumn{3}{c}{\textbf{00016}}                                                                                 & \multicolumn{3}{c}{\textbf{00019}}                                                                                 & \multicolumn{3}{c}{\textbf{00018}}                                                                                 & \multicolumn{4}{c}{\textbf{00027}}                                                                                                                       & \multicolumn{3}{c}{\textbf{All}}                                                \\ \midrule
Method                     & \multicolumn{1}{l}{\up{PSNR}} & \multicolumn{1}{l}{\up{SSIM}} & \multicolumn{1}{l}{\down{LPIPS}} & \multicolumn{1}{l}{\up{PSNR}} & \multicolumn{1}{l}{\up{SSIM}} & \multicolumn{1}{l}{\down{LPIPS}} & \multicolumn{1}{l}{\up{PSNR}} & \multicolumn{1}{l}{\up{SSIM}} & \multicolumn{1}{l}{\down{LPIPS}} & \multicolumn{1}{l}{\up{PSNR}} & \multicolumn{1}{l}{\up{SSIM}} & \multicolumn{1}{l}{\down{LPIPS}} & \multicolumn{1}{l}{\up{PSNR}} & \multicolumn{1}{l}{\up{SSIM}} & \multicolumn{1}{l}{\down{LPIPS}} &  \\
HAHA \cite{svitov2024haha} & 25.30                               & 0.9290                              & \best{0.0524}                                 & 27.66                               & 0.9576                              & \best{0.0501}                                 & 30.45                               & 0.9607                              & \best{0.0560}                                 & 27.38                               & 0.9510                              & \best{0.0479}                                 & 27.70                               & 0.9496                              & \best{0.0516}                                 &  \\
Skinned                    & \best{26.06}                               & \best{0.9299}                              & 0.0629                                 & \best{27.94}                               & \best{0.9605}                              & 0.0593                                 & \best{30.86}                               & \best{0.9621}                              & 0.0634                                 & \best{28.76}                               & \best{0.9553}                              & 0.0570                                 & \best{28.40}                               & \best{0.9519}                              & 0.0606                                 &  \\ \bottomrule
\end{tabular}%
}
\caption{Results on XHuman \cite{moon2024expressive}. We use the same subjects used by HAHA \cite{svitov2024haha} and compare on the test takes.}
\label{tab:xhuman}
\end{table*}

\subsection{A pure skinned gaussian avatar baseline}
\label{sec:sga_comparison}
We also performed an experiment on the XHuman dataset \cite{shen2023x} that uses the SMPL-X body model \cite{Pavlakos_2019_CVPR}, to create expressive avatars \cite{moon2024expressive}.
For this experiment, we design and implement a two-stage approach that uses no neural networks, serving as a baseline for fitting Gaussian avatars using color inputs, silhouettes, and parametric body fits.
For the first stage, we preprocess the body fits and silhouette images to deform the body mesh, aiming to initialize the Gaussian avatar fitting task with a higher-quality geometry.
For this stage, we use nvdiffrast \cite{laine2020modular} and use an intersection over union objective, supplemented with the mesh regularization objectives described in \cite{jena2023mesh}.
We model geometry deformation as a scalar offset $o$ per vertex $\boldsymbol{v}$, which deforms it along its normal direction $\boldsymbol{n}$, with the deformed position given by $\hat{\boldsymbol{x}} = \boldsymbol{x} + \boldsymbol{n} * o$.

When deforming the mesh by fitting these offsets, we subdivide the body mesh once.
For the next stage, we subdivide the deformed mesh once more to increase the density of the Gaussians $\mathcal{G}_i$, which are defined on the vertices and skinned following \S\ref{method:skinned}.
Apart from the rotation, SH coefficients, scale and opacity properties, we model position $\position = \hat{\boldsymbol{x}} + o' * \hat{\boldsymbol{n}}$ with an additional offset $o'$ along the normal directions, allowing the Gaussians to slightly rearrange their placement.
We optimize the skinned Gaussians using photometric (L1 and SSIM) and silhouette (L1) objectives by employing a modified Gaussian splat rasterizer that also outputs opacity as a mask.
We also use the long- and thin-scale regularization of SplattingAvatar \cite{shao2024splattingavatar}, in addition to L2 regularization for the scale and opacity.
We do not perform densification or pruning and use the default parameterized Adam optimizer \cite{kingma2014adam}.

\begin{figure*}[!htbp]
    \centering
    \begin{adjustbox}{width=\textwidth}
    \begin{tabular}{c}
        \begin{subfigure}{\textwidth}
            \centering
            \includegraphics[width=\textwidth]{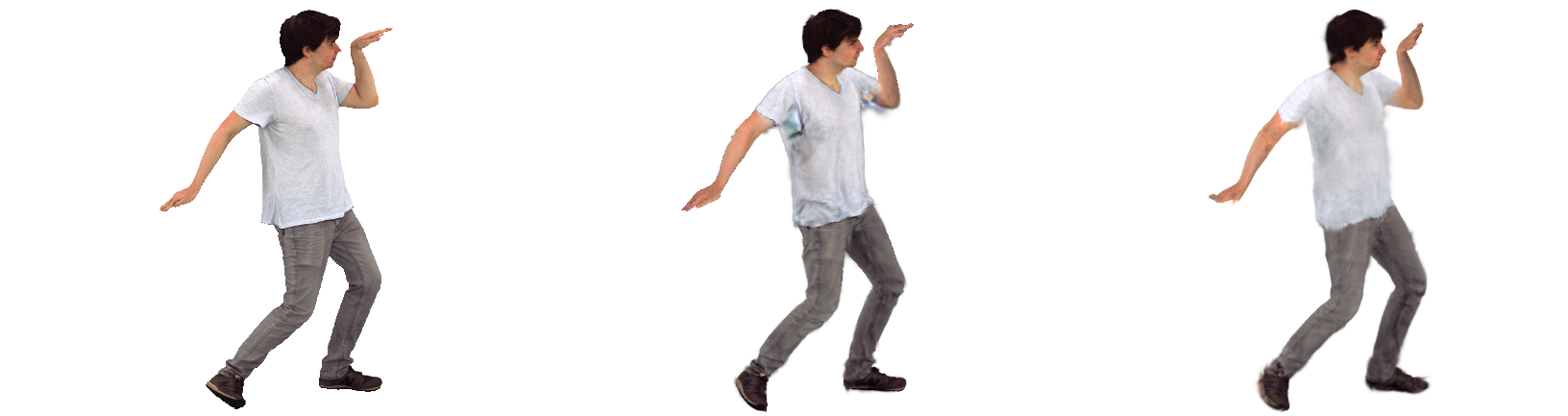}
        \end{subfigure} \\

        \begin{subfigure}{\textwidth}
            \centering
            \includegraphics[width=\textwidth]{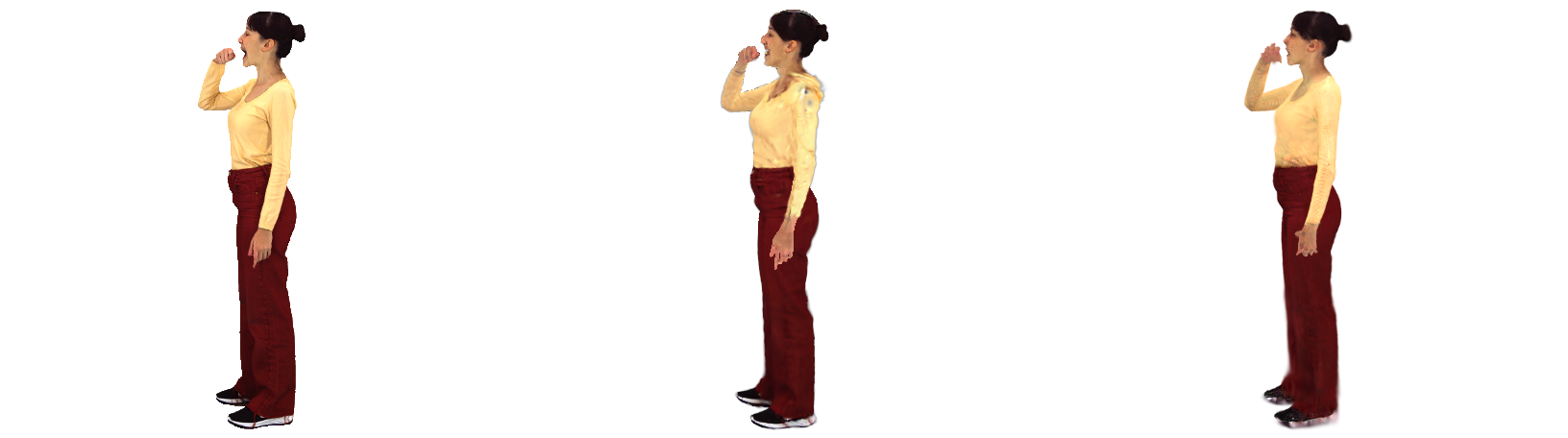}
        \end{subfigure} \\
        
        \begin{subfigure}{\textwidth}
            \centering
            \includegraphics[width=\textwidth]{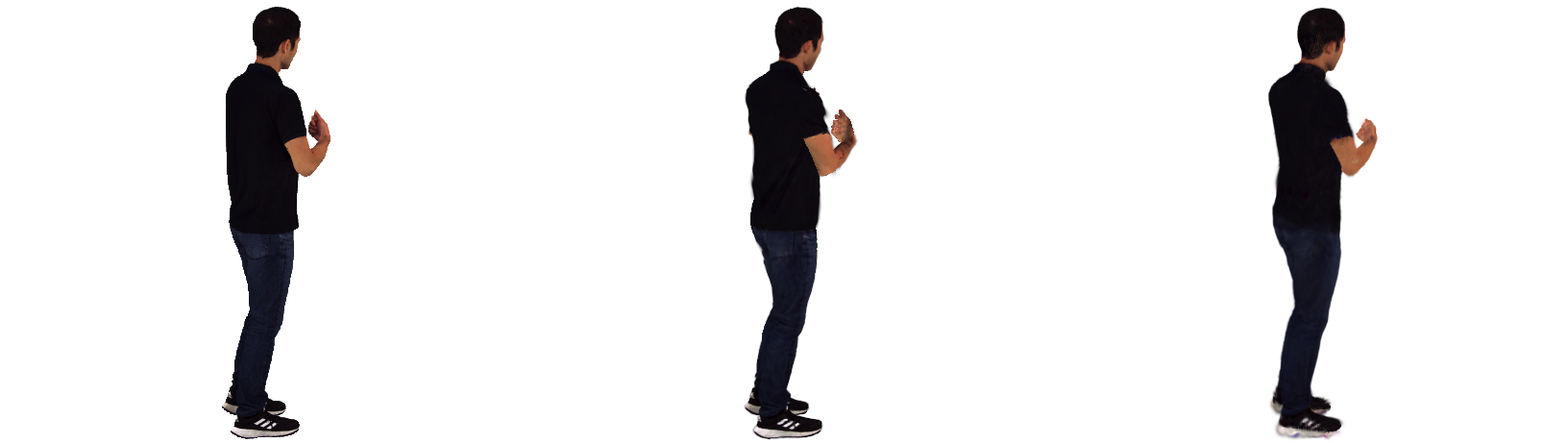}
        \end{subfigure} \\

        \begin{subfigure}{\textwidth}
            \centering
            \includegraphics[width=\textwidth]{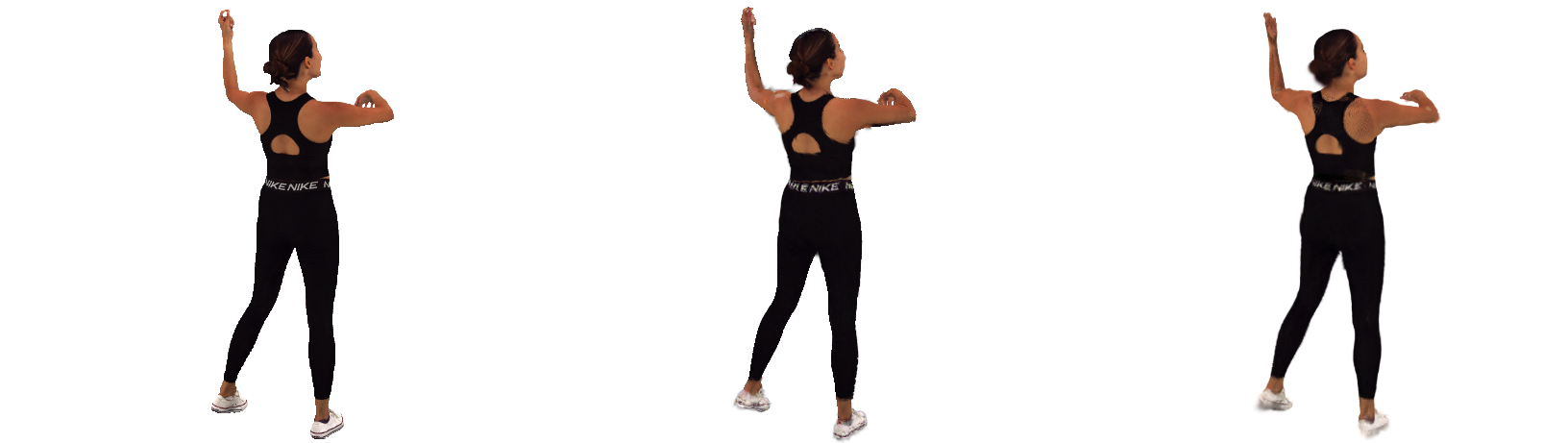}
        \end{subfigure} \\
    \end{tabular}
    \end{adjustbox}
    \caption{Qualitative comparison of our skinned baseline (right) with the results of HAHA \cite{svitov2024haha} (middle) and the ground-truth (left).}
    \label{fig:haha}
\end{figure*}

We compare it with a recent approach that combines Gaussian splats and traditional textured meshes in a hybrid representation, HAHA \cite{svitov2024haha}, one that also uses an explicit representation, without any neural networks.
For a fair comparison, we equalize the iterations performed over the data samples, using one-quarter of the iterations for the first mesh deformation stage and the rest of the iterations for the second stage, which optimizes the skinned Gaussians.
It should be noted that, compared to the previous datasets that contain rather simple motions, this dataset contains finger motion and more complex body motion, exhibiting more realistic avatar acquisition and simulation scenarios.
HAHA uses a single take for training, but crucially optimizes the SMPL-X poses to refine them, a step that has been shown to greatly improve performance. It also performs this refinement when testing on another take.
This overcomes issues related to image misalignment, which stabilizes the optimization process as the objectives and gradient directions are consistent.
We follow a different strategy to assess the effectiveness of our skinned Gaussians.
Instead of refining poses, we use more samples, essentially training on more unrefined takes.
We tested both approaches in the XHuman test split, and present the results for $4$ subjects in Table~\ref{tab:xhuman}.
We find that our pure Gaussian baseline exhibit more consistent visual quality performance, albeit at a higher number of splat levels and with less details preservation.
Our approach suffers on the LPIPS metric, which is more sensitive to image misalignments due to the lack of pose refinement.
Also, our approach uses skinned Gaussians that can be animated by only slightly modifying an existing LBS skinning implementation and then followed up by any Gaussian splat rasterizer, compared to custom multipass implementations for hybrid representations or Gaussian avatars that employ neural networks and custom processing.
Figure~\ref{fig:haha} presents a qualitative comparison of the skinned avatar baseline and the hybrid textured mesh and Gaussian splats avatars by HAHA. 
Evidently HAHA preserves details due to pose refinement, but this approach has not been scaled to a larger variety of poses that are necessary to produce highly articulated avatars.
Skinning Gaussian avatars allows for training with more poses, even unrefined ones. 
Nonetheless, pose errors -- especially at end effectors -- are detrimental to consistency, leading to missing fingers and blurrier results.
This shows that properly skinning the Gaussians is important as we progress towards building higher-quality animated avatars.

\subsection{In-the-wild skinned gaussian avatars}
\label{sec:wild_results}
To demonstrate the efficiency of the pure skinned Gaussian avatar, we implement an in-engine rendering pipeline in Unity3D.
Since our avatar is vertex-based, we adapt an LBS implementation using compute shaders to output rotated Gaussian properties.
We use a fixed number of $15$ iterations and the power iteration method for the eigenvalue decomposition of Eq.\eqref{average_quaternion} while skipping view-dependent effects.
Then we use an existing Gaussian rasterizer implementation\footnote{\href{https://github.com/aras-p/UnityGaussianSplatting}{https://github.com/aras-p/UnityGaussianSplatting}} to render the animated Gaussians.
This showcases the in-engine, plug-n-play nature of the skinned approach, which achieves over $200$ FPS rendering rates for $6$ simultaneous avatars on a laptop GPU (3080 Ti mobile).

\section{Conclusion}
Gaussian avatars are currently the most efficient representation for digitizing animated radiance fields of humans.
In addition to their rendering speed, they can also be optimized using forward skinning, which greatly improves their fitting process and runtime.
Up to now, linear blend skinning has been used to animate the positions, and sometimes the rotation and spherical harmonics components of Gaussian avatars.
However, linearly blended rotations are not proper rotation matrices, and this has been hurting their fitting process.
In this work we show that by properly blending the skinned rotations, we can both improve the Gaussians' fitting process, and also use the proper rotations to skin the spherical harmonics components.
This leads to improved Gaussian fitting, as demonstrated after integrating this Gaussian skinning technique to various methods in the literature.
We also provide an efficient, pure Gaussian baseline for reconstructing avatars and show that vertex attached Gaussians can be efficiently rendered in-engine.

{
    \small
    \bibliographystyle{ieeenat_fullname}
    \bibliography{main}

\begin{thebibliography}{76}
\providecommand{\natexlab}[1]{#1}
\providecommand{\url}[1]{\texttt{#1}}
\expandafter\ifx\csname urlstyle\endcsname\relax
  \providecommand{\doi}[1]{doi: #1}\else
  \providecommand{\doi}{doi: \begingroup \urlstyle{rm}\Url}\fi

\bibitem[Alldieck et~al.(2018{\natexlab{a}})Alldieck, Magnor, Xu, Theobalt, and Pons-Moll]{alldieck2018detailed}
Thiemo Alldieck, Marcus Magnor, Weipeng Xu, Christian Theobalt, and Gerard Pons-Moll.
\newblock Detailed human avatars from monocular video.
\newblock In \emph{2018 International Conference on 3D Vision (3DV)}, pages 98--109. IEEE, 2018{\natexlab{a}}.

\bibitem[Alldieck et~al.(2018{\natexlab{b}})Alldieck, Magnor, Xu, Theobalt, and Pons-Moll]{alldieck2018video}
Thiemo Alldieck, Marcus Magnor, Weipeng Xu, Christian Theobalt, and Gerard Pons-Moll.
\newblock Video based reconstruction of 3d people models.
\newblock In \emph{Proceedings of the IEEE Conference on Computer Vision and Pattern Recognition}, pages 8387--8397, 2018{\natexlab{b}}.

\bibitem[Aubert(2013)]{aubert2013alternative}
G Aubert.
\newblock An alternative to wigner d-matrices for rotating real spherical harmonics.
\newblock \emph{AIP advances}, 3\penalty0 (6), 2013.

\bibitem[Br{\'e}gier(2021)]{bregier2021deepregression}
Romain Br{\'e}gier.
\newblock Deep regression on manifolds: a {3D} rotation case study.
\newblock 2021.

\bibitem[Chen et~al.(2021{\natexlab{a}})Chen, Zhang, Kang, Zhe, Bao, Jia, and Lu]{chen2021animatable}
Jianchuan Chen, Ying Zhang, Di Kang, Xuefei Zhe, Linchao Bao, Xu Jia, and Huchuan Lu.
\newblock Animatable neural radiance fields from monocular rgb videos.
\newblock \emph{arXiv preprint arXiv:2106.13629}, 2021{\natexlab{a}}.

\bibitem[Chen et~al.(2024)Chen, Cong, and Liu]{chen2024saga}
Ronghan Chen, Yang Cong, and Jiayue Liu.
\newblock Saga: Surface-aligned gaussian avatar.
\newblock \emph{arXiv preprint arXiv:2412.00845}, 2024.

\bibitem[Chen et~al.(2021{\natexlab{b}})Chen, Zheng, Black, Hilliges, and Geiger]{chen2021snarf}
Xu Chen, Yufeng Zheng, Michael~J Black, Otmar Hilliges, and Andreas Geiger.
\newblock Snarf: Differentiable forward skinning for animating non-rigid neural implicit shapes.
\newblock In \emph{Proceedings of the IEEE/CVF International Conference on Computer Vision}, pages 11594--11604, 2021{\natexlab{b}}.

\bibitem[Chen et~al.(2023)Chen, Jiang, Song, Rietmann, Geiger, Black, and Hilliges]{chen2023fast}
Xu Chen, Tianjian Jiang, Jie Song, Max Rietmann, Andreas Geiger, Michael~J Black, and Otmar Hilliges.
\newblock Fast-snarf: A fast deformer for articulated neural fields.
\newblock \emph{IEEE Transactions on Pattern Analysis and Machine Intelligence}, 45\penalty0 (10):\penalty0 11796--11809, 2023.

\bibitem[Choutas et~al.(2020)Choutas, Pavlakos, Bolkart, Tzionas, and Black]{choutas2020monocular}
Vasileios Choutas, Georgios Pavlakos, Timo Bolkart, Dimitrios Tzionas, and Michael~J Black.
\newblock Monocular expressive body regression through body-driven attention.
\newblock In \emph{Computer Vision--ECCV 2020: 16th European Conference, Glasgow, UK, August 23--28, 2020, Proceedings, Part X 16}, pages 20--40. Springer, 2020.

\bibitem[Collet et~al.(2015)Collet, Chuang, Sweeney, Gillett, Evseev, Calabrese, Hoppe, Kirk, and Sullivan]{collet2015high}
Alvaro Collet, Ming Chuang, Pat Sweeney, Don Gillett, Dennis Evseev, David Calabrese, Hugues Hoppe, Adam Kirk, and Steve Sullivan.
\newblock High-quality streamable free-viewpoint video.
\newblock \emph{ACM Transactions on Graphics (ToG)}, 34\penalty0 (4):\penalty0 1--13, 2015.

\bibitem[Dong et~al.(2024)Dong, Xu, Gao, Bao, Xu, and Lau]{dong2024gaussian}
Zheng Dong, Ke Xu, Yaoan Gao, Hujun Bao, Weiwei Xu, and Rynson~WH Lau.
\newblock Gaussian surfel splatting for live human performance capture.
\newblock \emph{ACM Transactions on Graphics (TOG)}, 43\penalty0 (6):\penalty0 1--17, 2024.

\bibitem[Fei et~al.(2024)Fei, Xu, Zhang, Zhou, Yang, and He]{fei20243d}
Ben Fei, Jingyi Xu, Rui Zhang, Qingyuan Zhou, Weidong Yang, and Ying He.
\newblock 3d gaussian splatting as new era: A survey.
\newblock \emph{IEEE Transactions on Visualization and Computer Graphics}, 2024.

\bibitem[Gilmore(2008)]{gilmore2008lie}
Robert Gilmore.
\newblock \emph{Lie groups, physics, and geometry: an introduction for physicists, engineers and chemists}.
\newblock Cambridge University Press, 2008.

\bibitem[Goel et~al.(2023)Goel, Pavlakos, Rajasegaran, Kanazawa, and Malik]{goel2023humans}
Shubham Goel, Georgios Pavlakos, Jathushan Rajasegaran, Angjoo Kanazawa, and Jitendra Malik.
\newblock Humans in 4d: Reconstructing and tracking humans with transformers.
\newblock In \emph{Proceedings of the IEEE/CVF International Conference on Computer Vision}, pages 14783--14794, 2023.

\bibitem[Habermann et~al.(2021)Habermann, Liu, Xu, Zollhoefer, Pons-Moll, and Theobalt]{habermann2021real}
Marc Habermann, Lingjie Liu, Weipeng Xu, Michael Zollhoefer, Gerard Pons-Moll, and Christian Theobalt.
\newblock Real-time deep dynamic characters.
\newblock \emph{ACM Transactions on Graphics (ToG)}, 40\penalty0 (4):\penalty0 1--16, 2021.

\bibitem[Hu et~al.(2024{\natexlab{a}})Hu, Zhang, Zhang, Zhou, Liu, Zhang, and Nie]{hu2024gaussianavatar}
Liangxiao Hu, Hongwen Zhang, Yuxiang Zhang, Boyao Zhou, Boning Liu, Shengping Zhang, and Liqiang Nie.
\newblock Gaussianavatar: Towards realistic human avatar modeling from a single video via animatable 3d gaussians.
\newblock In \emph{Proceedings of the IEEE/CVF conference on computer vision and pattern recognition}, pages 634--644, 2024{\natexlab{a}}.

\bibitem[Hu et~al.(2024{\natexlab{b}})Hu, Hu, and Liu]{hu2024gauhuman}
Shoukang Hu, Tao Hu, and Ziwei Liu.
\newblock Gauhuman: Articulated gaussian splatting from monocular human videos.
\newblock In \emph{Proceedings of the IEEE/CVF conference on computer vision and pattern recognition}, pages 20418--20431, 2024{\natexlab{b}}.

\bibitem[Ichim et~al.(2015)Ichim, Bouaziz, and Pauly]{ichim2015dynamic}
Alexandru~Eugen Ichim, Sofien Bouaziz, and Mark Pauly.
\newblock Dynamic 3d avatar creation from hand-held video input.
\newblock \emph{ACM Transactions on Graphics (ToG)}, 34\penalty0 (4):\penalty0 1--14, 2015.

\bibitem[Jena et~al.(2023{\natexlab{a}})Jena, Chaudhari, Gee, Iyer, Choudhary, and Smith]{jena2023mesh}
Rohit Jena, Pratik Chaudhari, James Gee, Ganesh Iyer, Siddharth Choudhary, and Brandon~M Smith.
\newblock Mesh strikes back: Fast and efficient human reconstruction from rgb videos.
\newblock \emph{arXiv preprint arXiv:2303.08808}, 2023{\natexlab{a}}.

\bibitem[Jena et~al.(2023{\natexlab{b}})Jena, Iyer, Choudhary, Smith, Chaudhari, and Gee]{jena2023splatarmor}
Rohit Jena, Ganesh~Subramanian Iyer, Siddharth Choudhary, Brandon Smith, Pratik Chaudhari, and James Gee.
\newblock Splatarmor: Articulated gaussian splatting for animatable humans from monocular rgb videos.
\newblock \emph{arXiv preprint arXiv:2311.10812}, 2023{\natexlab{b}}.

\bibitem[Jiang et~al.(2022)Jiang, Yi, Samei, Tuzel, and Ranjan]{jiang2022neuman}
Wei Jiang, Kwang~Moo Yi, Golnoosh Samei, Oncel Tuzel, and Anurag Ranjan.
\newblock Neuman: Neural human radiance field from a single video.
\newblock In \emph{European Conference on Computer Vision}, pages 402--418. Springer, 2022.

\bibitem[Jung et~al.(2023)Jung, Brasch, Song, Perez-Pellitero, Zhou, Li, Navab, and Busam]{jung2023deformable}
HyunJun Jung, Nikolas Brasch, Jifei Song, Eduardo Perez-Pellitero, Yiren Zhou, Zhihao Li, Nassir Navab, and Benjamin Busam.
\newblock Deformable 3d gaussian splatting for animatable human avatars.
\newblock \emph{arXiv preprint arXiv:2312.15059}, 2023.

\bibitem[Kanazawa et~al.(2021)Kanazawa, Ross, Li, and Yang]{52306}
Angjoo Kanazawa, David Ross, Ruilong Li, and Shan Yang.
\newblock Ai choreographer: Music conditioned 3d dance generation with aist++.
\newblock 2021.

\bibitem[Kavan et~al.(2007)Kavan, Collins, {\v{Z}}{\'a}ra, and O'Sullivan]{kavan2007skinning}
Ladislav Kavan, Steven Collins, Ji{\v{r}}{\'\i} {\v{Z}}{\'a}ra, and Carol O'Sullivan.
\newblock Skinning with dual quaternions.
\newblock In \emph{Proceedings of the 2007 symposium on Interactive 3D graphics and games}, pages 39--46, 2007.

\bibitem[Ke et~al.(2022)Ke, Sun, Li, Yan, and Lau]{ke2022modnet}
Zhanghan Ke, Jiayu Sun, Kaican Li, Qiong Yan, and Rynson~WH Lau.
\newblock Modnet: Real-time trimap-free portrait matting via objective decomposition.
\newblock In \emph{Proceedings of the AAAI Conference on Artificial Intelligence}, pages 1140--1147, 2022.

\bibitem[Kerbl et~al.(2023)Kerbl, Kopanas, Leimk{\"u}hler, and Drettakis]{kerbl20233d}
Bernhard Kerbl, Georgios Kopanas, Thomas Leimk{\"u}hler, and George Drettakis.
\newblock 3d gaussian splatting for real-time radiance field rendering.
\newblock \emph{ACM Trans. Graph.}, 42\penalty0 (4):\penalty0 139--1, 2023.

\bibitem[Kingma and Ba(2014)]{kingma2014adam}
Diederik~P Kingma and Jimmy Ba.
\newblock Adam: A method for stochastic optimization.
\newblock \emph{arXiv preprint arXiv:1412.6980}, 2014.

\bibitem[Kocabas et~al.(2024)Kocabas, Chang, Gabriel, Tuzel, and Ranjan]{kocabas2024hugs}
Muhammed Kocabas, Jen-Hao~Rick Chang, James Gabriel, Oncel Tuzel, and Anurag Ranjan.
\newblock Hugs: Human gaussian splats.
\newblock In \emph{Proceedings of the IEEE/CVF conference on computer vision and pattern recognition}, pages 505--515, 2024.

\bibitem[Laine et~al.(2020)Laine, Hellsten, Karras, Seol, Lehtinen, and Aila]{laine2020modular}
Samuli Laine, Janne Hellsten, Tero Karras, Yeongho Seol, Jaakko Lehtinen, and Timo Aila.
\newblock Modular primitives for high-performance differentiable rendering.
\newblock \emph{ACM Transactions on Graphics (ToG)}, 39\penalty0 (6):\penalty0 1--14, 2020.

\bibitem[Lapchevskyi et~al.(2020)Lapchevskyi, Miller, Geiger, and Smidt]{lapchevskyi2020euclidean}
Kostiantyn Lapchevskyi, Benjamin Miller, Mario Geiger, and Tess Smidt.
\newblock Euclidean neural networks (e3nn) v1. 0.
\newblock Technical report, Lawrence Berkeley National Laboratory (LBNL), Berkeley, CA (United States), 2020.

\bibitem[Li et~al.(2023{\natexlab{a}})Li, Tao, Yang, and Yang]{li2023human101}
Mingwei Li, Jiachen Tao, Zongxin Yang, and Yi Yang.
\newblock Human101: Training 100+ fps human gaussians in 100s from 1 view.
\newblock \emph{arXiv preprint arXiv:2312.15258}, 2023{\natexlab{a}}.

\bibitem[Li et~al.(2023{\natexlab{b}})Li, Sun, Zheng, Wang, Zhang, and Liu]{li2023animatable}
Zhe Li, Yipengjing Sun, Zerong Zheng, Lizhen Wang, Shengping Zhang, and Yebin Liu.
\newblock Animatable and relightable gaussians for high-fidelity human avatar modeling.
\newblock \emph{arXiv preprint arXiv:2311.16096}, 2023{\natexlab{b}}.

\bibitem[Liang et~al.(2024{\natexlab{a}})Liang, Zhang, Feng, Shan, and Jia]{liang2024gs}
Zhihao Liang, Qi Zhang, Ying Feng, Ying Shan, and Kui Jia.
\newblock Gs-ir: 3d gaussian splatting for inverse rendering.
\newblock In \emph{Proceedings of the IEEE/CVF Conference on Computer Vision and Pattern Recognition}, pages 21644--21653, 2024{\natexlab{a}}.

\bibitem[Liang et~al.(2024{\natexlab{b}})Liang, Zhang, Hu, Zhu, Feng, and Jia]{liang2024analytic}
Zhihao Liang, Qi Zhang, Wenbo Hu, Lei Zhu, Ying Feng, and Kui Jia.
\newblock Analytic-splatting: Anti-aliased 3d gaussian splatting via analytic integration.
\newblock In \emph{European conference on computer vision}, pages 281--297. Springer, 2024{\natexlab{b}}.

\bibitem[Liao et~al.(2024)Liao, Golyanik, Habermann, and Theobalt]{liao2024vinecs}
Zhouyingcheng Liao, Vladislav Golyanik, Marc Habermann, and Christian Theobalt.
\newblock Vinecs: video-based neural character skinning.
\newblock In \emph{Proceedings of the IEEE/CVF Conference on Computer Vision and Pattern Recognition}, pages 1377--1387, 2024.

\bibitem[Liu et~al.(2021)Liu, Habermann, Rudnev, Sarkar, Gu, and Theobalt]{liu2021neural}
Lingjie Liu, Marc Habermann, Viktor Rudnev, Kripasindhu Sarkar, Jiatao Gu, and Christian Theobalt.
\newblock Neural actor: Neural free-view synthesis of human actors with pose control.
\newblock \emph{ACM transactions on graphics (TOG)}, 40\penalty0 (6):\penalty0 1--16, 2021.

\bibitem[Liu et~al.(2024)Liu, Huang, Qin, Lin, and Wang]{liu2024animatable}
Yang Liu, Xiang Huang, Minghan Qin, Qinwei Lin, and Haoqian Wang.
\newblock Animatable 3d gaussian: Fast and high-quality reconstruction of multiple human avatars.
\newblock In \emph{Proceedings of the 32nd ACM International Conference on Multimedia}, pages 1120--1129, 2024.

\bibitem[Loper et~al.(2015)Loper, Mahmood, Romero, Pons-Moll, and Black]{loper2015smpl}
Matthew Loper, Naureen Mahmood, Javier Romero, Gerard Pons-Moll, and Michael~J. Black.
\newblock Smpl: A skinned multi-person linear model.
\newblock \emph{ACM Transactions on Graphics (SIGGRAPH)}, 2015.

\bibitem[Luo et~al.(2025)Luo, Peng, Cai, Yang, Yang, Cao, and Lin]{luo2025deblur}
Xianrui Luo, Juewen Peng, Zhongang Cai, Lei Yang, Fan Yang, Zhiguo Cao, and Guosheng Lin.
\newblock Deblur-avatar: Animatable avatars from motion-blurred monocular videos.
\newblock \emph{arXiv preprint arXiv:2501.13335}, 2025.

\bibitem[Markley et~al.(2007)Markley, Cheng, Crassidis, and Oshman]{markley2007averaging}
F~Landis Markley, Yang Cheng, John~L Crassidis, and Yaakov Oshman.
\newblock Averaging quaternions.
\newblock \emph{Journal of Guidance, Control, and Dynamics}, 30\penalty0 (4):\penalty0 1193--1197, 2007.

\bibitem[Mildenhall et~al.(2021)Mildenhall, Srinivasan, Tancik, Barron, Ramamoorthi, and Ng]{mildenhall2021nerf}
Ben Mildenhall, Pratul~P Srinivasan, Matthew Tancik, Jonathan~T Barron, Ravi Ramamoorthi, and Ren Ng.
\newblock Nerf: Representing scenes as neural radiance fields for view synthesis.
\newblock \emph{Communications of the ACM}, 65\penalty0 (1):\penalty0 99--106, 2021.

\bibitem[Moon et~al.(2024)Moon, Shiratori, and Saito]{moon2024expressive}
Gyeongsik Moon, Takaaki Shiratori, and Shunsuke Saito.
\newblock Expressive whole-body 3d gaussian avatar.
\newblock In \emph{European Conference on Computer Vision}, pages 19--35. Springer, 2024.

\bibitem[Moreau et~al.(2024)Moreau, Song, Dhamo, Shaw, Zhou, and P{\'e}rez-Pellitero]{moreau2024human}
Arthur Moreau, Jifei Song, Helisa Dhamo, Richard Shaw, Yiren Zhou, and Eduardo P{\'e}rez-Pellitero.
\newblock Human gaussian splatting: Real-time rendering of animatable avatars.
\newblock In \emph{Proceedings of the IEEE/CVF conference on computer vision and pattern recognition}, pages 788--798, 2024.

\bibitem[Natsume et~al.(2019)Natsume, Saito, Huang, Chen, Ma, Li, and Morishima]{natsume2019siclope}
Ryota Natsume, Shunsuke Saito, Zeng Huang, Weikai Chen, Chongyang Ma, Hao Li, and Shigeo Morishima.
\newblock Siclope: Silhouette-based clothed people.
\newblock In \emph{Proceedings of the IEEE/CVF Conference on Computer Vision and Pattern Recognition}, pages 4480--4490, 2019.

\bibitem[Newcombe et~al.(2015)Newcombe, Fox, and Seitz]{newcombe2015dynamicfusion}
Richard~A Newcombe, Dieter Fox, and Steven~M Seitz.
\newblock Dynamicfusion: Reconstruction and tracking of non-rigid scenes in real-time.
\newblock In \emph{Proceedings of the IEEE conference on computer vision and pattern recognition}, pages 343--352, 2015.

\bibitem[Niu et~al.(2024)Niu, Zhan, Zhu, Li, Wang, Zhong, Sun, and Zheng]{niu2024bundle}
Muyao Niu, Yifan Zhan, Qingtian Zhu, Zhuoxiao Li, Wei Wang, Zhihang Zhong, Xiao Sun, and Yinqiang Zheng.
\newblock Bundle adjusted gaussian avatars deblurring.
\newblock \emph{arXiv preprint arXiv:2411.16758}, 2024.

\bibitem[Osman et~al.(2020)Osman, Bolkart, and Black]{osman2020star}
Ahmed A.~A. Osman, Timo Bolkart, and Michael~J. Black.
\newblock Star: Sparse trained articulated human body regressor.
\newblock \emph{European Conference on Computer Vision (ECCV)}, 2020.

\bibitem[Pang et~al.(2024)Pang, Zhu, Kortylewski, Theobalt, and Habermann]{pang2024ash}
Haokai Pang, Heming Zhu, Adam Kortylewski, Christian Theobalt, and Marc Habermann.
\newblock Ash: Animatable gaussian splats for efficient and photoreal human rendering.
\newblock In \emph{Proceedings of the IEEE/CVF Conference on Computer Vision and Pattern Recognition}, pages 1165--1175, 2024.

\bibitem[Paudel et~al.(2024)Paudel, Khanal, Paudel, Tandukar, and Chhatkuli]{paudel2024ihuman}
Pramish Paudel, Anubhav Khanal, Danda~Pani Paudel, Jyoti Tandukar, and Ajad Chhatkuli.
\newblock ihuman: Instant animatable digital humans from monocular videos.
\newblock In \emph{European Conference on Computer Vision}, pages 304--323. Springer, 2024.

\bibitem[Pavlakos et~al.(2019)Pavlakos, Choutas, Ghorbani, Bolkart, Osman, Tzionas, and Black]{Pavlakos_2019_CVPR}
Georgios Pavlakos, Vasileios Choutas, Nima Ghorbani, Timo Bolkart, Ahmed A.~A. Osman, Dimitrios Tzionas, and Michael~J. Black.
\newblock Expressive body capture: 3d hands, face, and body from a single image.
\newblock In \emph{Proceedings of the IEEE/CVF Conference on Computer Vision and Pattern Recognition (CVPR)}, 2019.

\bibitem[Peng et~al.(2021{\natexlab{a}})Peng, Dong, Wang, Zhang, Shuai, Zhou, and Bao]{peng2021animatable}
Sida Peng, Junting Dong, Qianqian Wang, Shangzhan Zhang, Qing Shuai, Xiaowei Zhou, and Hujun Bao.
\newblock Animatable neural radiance fields for modeling dynamic human bodies.
\newblock In \emph{Proceedings of the IEEE/CVF International Conference on Computer Vision}, pages 14314--14323, 2021{\natexlab{a}}.

\bibitem[Peng et~al.(2021{\natexlab{b}})Peng, Zhang, Xu, Wang, Shuai, Bao, and Zhou]{peng2021neural}
Sida Peng, Yuanqing Zhang, Yinghao Xu, Qianqian Wang, Qing Shuai, Hujun Bao, and Xiaowei Zhou.
\newblock Neural body: Implicit neural representations with structured latent codes for novel view synthesis of dynamic humans.
\newblock In \emph{Proceedings of the IEEE/CVF conference on computer vision and pattern recognition}, pages 9054--9063, 2021{\natexlab{b}}.

\bibitem[Qian et~al.(2024)Qian, Wang, Mihajlovic, Geiger, and Tang]{qian20243dgs}
Zhiyin Qian, Shaofei Wang, Marko Mihajlovic, Andreas Geiger, and Siyu Tang.
\newblock 3dgs-avatar: Animatable avatars via deformable 3d gaussian splatting.
\newblock In \emph{Proceedings of the IEEE/CVF conference on computer vision and pattern recognition}, pages 5020--5030, 2024.

\bibitem[Saito et~al.(2020)Saito, Simon, Saragih, and Joo]{saito2020pifuhd}
Shunsuke Saito, Tomas Simon, Jason Saragih, and Hanbyul Joo.
\newblock Pifuhd: Multi-level pixel-aligned implicit function for high-resolution 3d human digitization.
\newblock In \emph{Proceedings of the IEEE/CVF conference on computer vision and pattern recognition}, pages 84--93, 2020.

\bibitem[Shao et~al.(2024)Shao, Wang, Li, Wang, Lin, Zhang, Fan, and Wang]{shao2024splattingavatar}
Zhijing Shao, Zhaolong Wang, Zhuang Li, Duotun Wang, Xiangru Lin, Yu Zhang, Mingming Fan, and Zeyu Wang.
\newblock Splattingavatar: Realistic real-time human avatars with mesh-embedded gaussian splatting.
\newblock In \emph{Proceedings of the IEEE/CVF Conference on Computer Vision and Pattern Recognition}, pages 1606--1616, 2024.

\bibitem[Shen et~al.(2023)Shen, Guo, Kaufmann, Zarate, Valentin, Song, and Hilliges]{shen2023x}
Kaiyue Shen, Chen Guo, Manuel Kaufmann, Juan~Jose Zarate, Julien Valentin, Jie Song, and Otmar Hilliges.
\newblock X-avatar: Expressive human avatars.
\newblock In \emph{Proceedings of the IEEE/CVF Conference on Computer Vision and Pattern Recognition}, pages 16911--16921, 2023.

\bibitem[Su et~al.(2021)Su, Yu, Zollh{\"o}fer, and Rhodin]{su2021nerf}
Shih-Yang Su, Frank Yu, Michael Zollh{\"o}fer, and Helge Rhodin.
\newblock A-nerf: Articulated neural radiance fields for learning human shape, appearance, and pose.
\newblock \emph{Advances in neural information processing systems}, 34:\penalty0 12278--12291, 2021.

\bibitem[Svitov et~al.(2024)Svitov, Morerio, Agapito, and Del~Bue]{svitov2024haha}
David Svitov, Pietro Morerio, Lourdes Agapito, and Alessio Del~Bue.
\newblock Haha: Highly articulated gaussian human avatars with textured mesh prior.
\newblock In \emph{Proceedings of the Asian Conference on Computer Vision}, pages 4051--4068, 2024.

\bibitem[Talegaonkar et~al.(2024)Talegaonkar, Belhe, Ramamoorthi, and Antipa]{talegaonkar2024volumetrically}
Chinmay Talegaonkar, Yash Belhe, Ravi Ramamoorthi, and Nicholas Antipa.
\newblock Volumetrically consistent 3d gaussian rasterization.
\newblock \emph{arXiv preprint arXiv:2412.03378}, 2024.

\bibitem[Wang et~al.(2022{\natexlab{a}})Wang, Peng, Zhou, Yang, and Zhang]{wang2022nerfcap}
Kangkan Wang, Sida Peng, Xiaowei Zhou, Jian Yang, and Guofeng Zhang.
\newblock Nerfcap: Human performance capture with dynamic neural radiance fields.
\newblock \emph{IEEE Transactions on Visualization and Computer Graphics}, 29\penalty0 (12):\penalty0 5097--5110, 2022{\natexlab{a}}.

\bibitem[Wang et~al.(2024)Wang, Cao, Han, and Wong]{wang2024survey}
Ruihe Wang, Yukang Cao, Kai Han, and Kwan-Yee~K Wong.
\newblock A survey on 3d human avatar modeling--from reconstruction to generation.
\newblock \emph{arXiv preprint arXiv:2406.04253}, 2024.

\bibitem[Wang et~al.(2022{\natexlab{b}})Wang, Schwarz, Geiger, and Tang]{wang2022arah}
Shaofei Wang, Katja Schwarz, Andreas Geiger, and Siyu Tang.
\newblock Arah: Animatable volume rendering of articulated human sdfs.
\newblock In \emph{European conference on computer vision}, pages 1--19. Springer, 2022{\natexlab{b}}.

\bibitem[Weng et~al.(2022)Weng, Curless, Srinivasan, Barron, and Kemelmacher-Shlizerman]{weng2022humannerf}
Chung-Yi Weng, Brian Curless, Pratul~P Srinivasan, Jonathan~T Barron, and Ira Kemelmacher-Shlizerman.
\newblock Humannerf: Free-viewpoint rendering of moving people from monocular video.
\newblock In \emph{Proceedings of the IEEE/CVF conference on computer vision and pattern Recognition}, pages 16210--16220, 2022.

\bibitem[Wenninger et~al.(2020)Wenninger, Achenbach, Bartl, Latoschik, and Botsch]{wenninger2020realistic}
Stephan Wenninger, Jascha Achenbach, Andrea Bartl, Marc~Erich Latoschik, and Mario Botsch.
\newblock Realistic virtual humans from smartphone videos.
\newblock In \emph{Proceedings of the 26th ACM symposium on virtual reality software and technology}, pages 1--11, 2020.

\bibitem[Wigner(1931)]{wigner1931gruppentheorie}
Eugen Wigner.
\newblock Gruppentheorie und ihre anwendung auf die quantenmechanik der atomspektren.
\newblock 1931.

\bibitem[Xu et~al.(2021)Xu, Alldieck, and Sminchisescu]{xu2021h}
Hongyi Xu, Thiemo Alldieck, and Cristian Sminchisescu.
\newblock H-nerf: Neural radiance fields for rendering and temporal reconstruction of humans in motion.
\newblock \emph{Advances in Neural Information Processing Systems}, 34:\penalty0 14955--14966, 2021.

\bibitem[Yan et~al.(2024)Yan, Low, Chen, and Lee]{yan2024multi}
Zhiwen Yan, Weng~Fei Low, Yu Chen, and Gim~Hee Lee.
\newblock Multi-scale 3d gaussian splatting for anti-aliased rendering.
\newblock In \emph{Proceedings of the IEEE/CVF Conference on Computer Vision and Pattern Recognition}, pages 20923--20931, 2024.

\bibitem[Yu et~al.(2018)Yu, Zheng, Guo, Zhao, Dai, Li, Pons-Moll, and Liu]{yu2018doublefusion}
Tao Yu, Zerong Zheng, Kaiwen Guo, Jianhui Zhao, Qionghai Dai, Hao Li, Gerard Pons-Moll, and Yebin Liu.
\newblock Doublefusion: Real-time capture of human performances with inner body shapes from a single depth sensor.
\newblock In \emph{Proceedings of the IEEE conference on computer vision and pattern recognition}, pages 7287--7296, 2018.

\bibitem[Zhang et~al.(2023)Zhang, Tian, Zhang, Li, An, Sun, and Liu]{zhang2023pymaf}
Hongwen Zhang, Yating Tian, Yuxiang Zhang, Mengcheng Li, Liang An, Zhenan Sun, and Yebin Liu.
\newblock Pymaf-x: Towards well-aligned full-body model regression from monocular images.
\newblock \emph{IEEE Transactions on Pattern Analysis and Machine Intelligence}, 45\penalty0 (10):\penalty0 12287--12303, 2023.

\bibitem[Zhang et~al.(2018)Zhang, Isola, Efros, Shechtman, and Wang]{zhang2018unreasonable}
Richard Zhang, Phillip Isola, Alexei~A Efros, Eli Shechtman, and Oliver Wang.
\newblock The unreasonable effectiveness of deep features as a perceptual metric.
\newblock In \emph{Proceedings of the IEEE conference on computer vision and pattern recognition}, pages 586--595, 2018.

\bibitem[Zhao et~al.(2022)Zhao, Yang, Zhang, Lin, Zhang, Yu, and Xu]{zhao2022humannerf}
Fuqiang Zhao, Wei Yang, Jiakai Zhang, Pei Lin, Yingliang Zhang, Jingyi Yu, and Lan Xu.
\newblock Humannerf: Efficiently generated human radiance field from sparse inputs.
\newblock In \emph{Proceedings of the IEEE/CVF Conference on Computer Vision and Pattern Recognition}, pages 7743--7753, 2022.

\bibitem[Zhi et~al.(2020)Zhi, Lassner, Tung, Stoll, Narasimhan, and Vo]{zhi2020texmesh}
Tiancheng Zhi, Christoph Lassner, Tony Tung, Carsten Stoll, Srinivasa~G Narasimhan, and Minh Vo.
\newblock Texmesh: Reconstructing detailed human texture and geometry from rgb-d video.
\newblock In \emph{Computer Vision--ECCV 2020: 16th European Conference, Glasgow, UK, August 23--28, 2020, Proceedings, Part X 16}, pages 492--509. Springer, 2020.

\bibitem[Zhou et~al.(2019)Zhou, Barnes, Lu, Yang, and Li]{zhou2019continuity}
Yi Zhou, Connelly Barnes, Jingwan Lu, Jimei Yang, and Hao Li.
\newblock On the continuity of rotation representations in neural networks.
\newblock In \emph{Proceedings of the IEEE/CVF conference on computer vision and pattern recognition}, pages 5745--5753, 2019.

\bibitem[Zielonka et~al.(2023)Zielonka, Bagautdinov, Saito, Zollh{\"o}fer, Thies, and Romero]{zielonka2023drivable}
Wojciech Zielonka, Timur Bagautdinov, Shunsuke Saito, Michael Zollh{\"o}fer, Justus Thies, and Javier Romero.
\newblock Drivable 3d gaussian avatars.
\newblock \emph{arXiv preprint arXiv:2311.08581}, 2023.

\bibitem[Zuo et~al.(2020)Zuo, Wang, Zheng, Yu, Gong, Yang, and Cheng]{zuo2020sparsefusion}
Xinxin Zuo, Sen Wang, Jiangbin Zheng, Weiwei Yu, Minglun Gong, Ruigang Yang, and Li Cheng.
\newblock Sparsefusion: Dynamic human avatar modeling from sparse rgbd images.
\newblock \emph{IEEE Transactions on Multimedia}, 23:\penalty0 1617--1629, 2020.

\bibitem[Zwicker et~al.(2002)Zwicker, Pfister, Van~Baar, and Gross]{zwicker2002ewa}
Matthias Zwicker, Hanspeter Pfister, Jeroen Van~Baar, and Markus Gross.
\newblock Ewa splatting.
\newblock \emph{IEEE Transactions on Visualization and Computer Graphics}, 8\penalty0 (3):\penalty0 223--238, 2002.

\end{thebibliography}
}

\end{document}